\documentclass[conference]{IEEEtran}
\IEEEoverridecommandlockouts

\usepackage{cite}
\usepackage{amsmath,amssymb,amsfonts}
\usepackage{graphicx}
\usepackage[bookmarks=false]{hyperref}
\usepackage{url}
\usepackage[labelformat=simple]{subcaption}

\begin{document}

\title{Explainable Deep CNNs for MRI-Based Diagnosis of Alzheimer's Disease\\
\thanks{We thank the partial support given by the Project: Models, Algorithms and Systems for the Web (grant FAPEMIG / PRONEX / MASWeb APQ-01400-14), and authors' individual grants and scholarships from CNPq, Fapemig and Kunumi.}
}

\author{\IEEEauthorblockN{Eduardo Nigri, Nivio Ziviani}
\IEEEauthorblockA{
\textit{CS Dept. UFMG \& Kunumi}\\
Belo Horizonte\\
Brazil\\
\{eduardonigri, nivio\}@dcc.ufmg.br}
\and
\IEEEauthorblockN{Fabio Cappabianco}
\IEEEauthorblockA{
\textit{UNIFESP DCT}\\
S\~{a}o Jos\'{e} dos Campos\\
Brazil\\
cappabianco@unifesp.br}
\and
\IEEEauthorblockN{Augusto Antunes}
\IEEEauthorblockA{
\textit{InRad-FMUSP \& Kunumi}\\
S\~{a}o Paulo\\
Brazil\\
augustobfantunes@gmail.com}
\and
\IEEEauthorblockN{Adriano Veloso}
\IEEEauthorblockA{
\textit{CS Dept. UFMG}\\
Belo Horizonte\\
Brazil\\
adrianov@dcc.ufmg.br}
\and
\IEEEauthorblockN{~~~~}
\IEEEauthorblockA{}
\and
\IEEEauthorblockN{~~~~~~~~~~~~~~~~~~~~~~~~~~~~~~~~~~~~the Alzheimer's Disease Neuroimaging Initiative*\thanks{*Data used in preparation of this article were obtained from the Alzheimer's Disease Neuroimaging Initiative (ADNI) database (adni.loni.usc.edu). As such, the investigators within the ADNI contributed to the design and implementation of ADNI and/or provided data but did not participate in analysis or writing of this report. A complete listing of ADNI investigators can be found at: http://adni.loni.usc.edu/wp-content/uploads/how\_to\_apply/ADNI\_Acknowledgement\_List.pdf}}
}

\maketitle

\begin{abstract}
Deep Convolutional Neural Networks (CNNs) are becoming prominent models for semi-automated diagnosis of Alzheimer's Disease (AD) using brain Magnetic Resonance Imaging (MRI). Although being highly accurate, deep CNN models lack transparency and interpretability, precluding adequate clinical reasoning and not complying with most current regulatory demands. One popular choice for explaining deep image models is occluding regions of the image to isolate their influence on the prediction. However, existing methods for occluding patches of brain scans generate images outside the distribution to which the model was trained for, thus leading to unreliable explanations. In this paper, we propose an alternative explanation method that is specifically designed for the brain scan task. Our method, which we refer to as Swap Test, produces heatmaps that depict the areas of the brain that are most indicative of AD, providing interpretability for the model's decisions in a format understandable to clinicians. Experimental results using an axiomatic evaluation show that the proposed method is more suitable for explaining the diagnosis of AD using MRI while the opposite trend was observed when using a typical occlusion test. Therefore, we believe our method may address the inherent black-box nature of deep neural networks that are capable of diagnosing AD.
\end{abstract}


\section{Introduction}
\label{sec:introduction}

Alzheimer's Disease (AD) is a neurodegenerative disease that is widely considered to be one of the major future challenges in healthcare. It is currently the sixth-leading cause of death and affects about 5.7 million people in the U.S. alone, with estimates to increase to 14 million by 2050 \cite{alz2018}. The disease accounts for 60\% to 80\% of all the dementia cases, affecting mainly individuals over 65 years old \cite{Pouryamout2012}. The symptoms include memory loss, impaired reasoning or judgment, mood and behavior changes, disorientation, and difficulties with language. The pathological hallmark of AD is the accumulation of beta-amyloid plaques and neurofibrillary tangles in specific regions of the brain, particularly the hippocampus \cite{Richard2017}. There is currently no cure for the disease, and early diagnosis is crucial to preventive measures.

The diagnosis of the disease can be made using different methods, such as lumbar puncture \cite{Shaw@AnnalsofNeurology2009}, blood tests \cite{Shi2018}, structural Magnetic Resonance Imaging (MRI) \cite{Sorensen2017}, functional MRI \cite{Sarraf2016}, and Positron Emission Tomography \cite{Mathotaarachchi2017}. While each type of exam has its advantages, structural MRI is desirable for its low invasiveness and sensibility to early changes in the affected regions of the brain \cite{Frisoni2010}.

An automated diagnosis of AD could be highly beneficial to patients and consequently, the use of Machine Learning using MRI data has been a very active area of research. The most common approach is to classify subjects as either Cognitively Normal (CN) or AD. Earlier studies have used handcrafted features combined with general classifiers \cite{Klein2010,Lillemark@BMCMedicalImaging2014}. Most of these models use measurements of relevant brain structures such as hippocampal and ventricular volume as features. In recent years, the use of deep learning architectures for automated feature learning has been the method of choice. In particular, several studies have applied stacked auto-encoders \cite{Liu2014,Dolph@IJCNN2017} and 3D convolutional networks \cite{payan2015,Backstrom@ISBI2018,Wegmayr@SPIEMedicalImaging2018} or variations of both.

While the aforementioned deep models have consistently achieved high accuracy figures in distinguishing between CN and AD subjects $-$ in many cases, on par with the performance of experienced clinicians~\cite{Backstrom@ISBI2018,Wegmayr@SPIEMedicalImaging2018} $-$ it is still unclear if the high accuracy measured results from appropriate problem modeling rather than the exploitation of correlative factors present in the data. Since deep models are intrinsically complex, it is not clear what information in the brain scan makes the model actually arrive at its decisions. As mistakes can have catastrophic effects, this black-box aspect of deep models makes it difficult for doctors to trust them. In this sense, there are many arguments in favor of model explainability while separating CN from AD subjects, including:

\begin{itemize}
\item Validation: Every model decision should be made clear for appropriate validation by a clinician so that the reliance of the model on the correct features is guaranteed. Validation is particularly important to assure that the model is not exploiting artifacts in the data. With deep models, however, the challenge for clinicians is that they have no context for why a diagnosis was chosen.
\item Exploration: Explainable models trained on a large number of examples may provide insights about AD diagnosis, which were previously unclear to clinicians. Model explainability is thus paramount in view of the importance of anatomy in the interpretation of AD brains.
\item Compliance with regulations: New regulations propose that individuals affected by algorithmic decisions have a right to explanation~\cite{explanation}. Therefore, models will necessarily have to become explainable in order to provide satisfactory answers for legal questions.
\end{itemize}

Model explanation methods often rely on occluding regions of the image in order to isolate their influence on the model prediction \cite{Zeiler@ECCV2014}. This is performed by temporarily setting all pixels of this region to a particular value. Our hypothesis was that existing occlusion-based methods are ill-suited to the task of explaining AD diagnosis in deep models. This is because occluding patches of the image interferes with the predicted image label. 

Occlusion-based methods were proposed for image classification tasks in which the objects of interest are presented against a variety of backgrounds and color intensities. By contrast, in brain scans, the images are very similar and are normally registered so that the same brain regions occupy roughly the same positions. Occluding patches of the image generates images outside of the distribution to which the model was trained for, leading to unreliable explanations. For instance, occluding parts of a healthy brain would reduce the amount of brain tissue, a process that is also caused by the disease, and could effectively change the label of the image being explained.

In this paper, we propose an approach for explaining model decisions~\cite{ref10,ref12}, in the context of registered brain images, that addresses the issues of current occlusion-based methods. The method, which we refer to as Swap Test, produces a heatmap of the most relevant image regions according to a model for a single image. That is, our method provides a visual explanation as to why the diagnosis decision was made by highlighting regions of the brain scan, a format that can be easily understood by clinicians. We use an axiomatic approach to evaluate our hypothesis and our numerical results show that the proposed method is indeed superior as compared to the occlusion-based approach. This result is demonstrated using a representative range of CNN architectures, which have accurate predictions in comparison to a baseline.

The remainder of this paper is organized as follows. Section~2 discusses relevant related work. Section~3 presents our method for explainable AD diagnosis using deep learning models. In Section~4, we present our experimental setup, and in Section~5, we report our results. Finally, Section~6 concludes our paper.

\section{Related Work}

Convolutional Neural Network (CNN) architectures have been widely used in image recognition tasks since its introduction in \cite{LeCun1998} and superior performance on ImageNet in \cite{Krizhevsky@NIPS2012}. In more recent years, they also have been successfully applied to the area of medical image analysis, despite the additional challenges such as higher image complexity and lower amount of labeled data. Examples of such applications include brain tumor segmentation \cite{Isin2016}, diagnosis of diabetic retinopathy \cite{Rajalakshmi2018}, lung cancer nodule detection \cite{Song2017}, and pneumonia diagnosis from X-ray scans \cite{Rajpurkar2017}.

The problem of diagnosis of AD from MRI scans has been extensively studied and several methods have been proposed. The most common approach is to model the diagnosis as a classification problem, while approaches relying on cognitive tests also have been studied~\cite{Zhang2017,Zhang2017_2,Stonnington2010}. Methods based on handcrafted features and off-the-shelf classifiers~\cite{Lillemark@BMCMedicalImaging2014,Sorensen2017,Chincarini2011,Klein2010} frequently use hippocampal and cortical thickness measurements or brain similarity between subjects. Most of the recent studies used deep learning models, in particular, Stacked Auto-Encoders~\cite{Liu2014,payan2015,Dolph@IJCNN2017} and 3D CNNs \cite{payan2015, HosseiniAsl2016, Korolev2017, Wegmayr@SPIEMedicalImaging2018, Backstrom@ISBI2018} have been explored multiple times.

The work of Payan and colleagues~\cite{payan2015} was one of the first to apply 3D CNNs to this problem. A comparison with a 2D approach was also provided, with the 3D approach providing better results. The proposed 3-layer network was pre-trained using 3D sparse autoencoders to learn the convolutional features, which were then used as the first layers of the CNN. Hosseini-Asl and colleagues \cite{HosseiniAsl2016} proposed a deeply supervised adaptable 3D-CNN which uses 3D convolutional autoencoders (3D-CAE) as pre-training with domain adaptation. The 3D-CAE was trained using the source domain data extracted from CADDementia \cite{caddementia} and the 6-layer 3D-CNN was fine-tuned to a specific task using data from ADNI. 

Korolev and colleagues \cite{Korolev2017} compared two approaches for the binary classification problem, the former being inspired by VGG \cite{Simonyan2014} and the latter by Residual Neural Networks \cite{he2015}. The models were first adapted to operate in three dimensions. Both approaches were evaluated using data from ADNI that was preprocessed for alignment and skull-stripping. Wegmayr and colleagues \cite{Wegmayr@SPIEMedicalImaging2018} also applied a 3D-CNN, but with a larger dataset and a 7-layer network. They also identified several pitfalls in previous studies that, if not taken into account, could result in biased accuracy estimates. And finally, Bäckström and colleagues \cite{Backstrom@ISBI2018} also proposed a 3D-CNN, which contains 8 layers. They also studied the impact of hyperparameter selection, data preprocessing, dataset partitioning and dataset size.

While the previously mentioned studies achieved high accuracy when distinguishing between CN and AD, their use in real-life applications is still limited due to the lack of explainability of the classification decisions, which is a limitation that we propose to address in this paper.

\begin{figure*}
\centering
\includegraphics[width=0.9\linewidth]{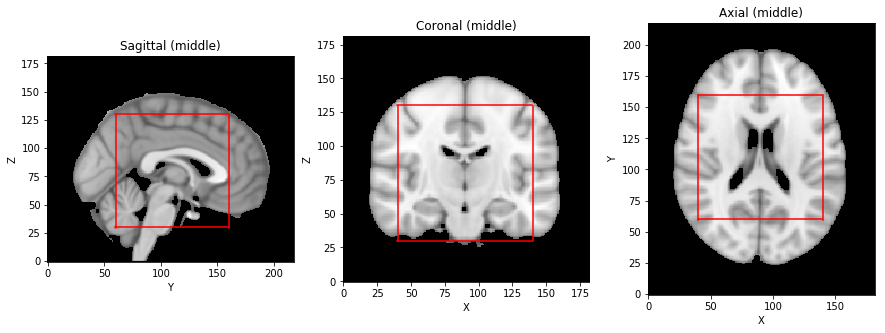}
\caption{Brain region extracted after the preprocessing steps superposed on the registration atlas.}
\label{brain_region}
\end{figure*}

\section{Method}
\label{sec:methodology}

Our methodology is divided into three steps: image preprocessing, classification models, and model explanation. This section explains each step in detail.

\subsection{Image Preprocessing}
\label{sec:imagepreprocessing}

The first step prepares the raw MRI images by removing unnecessary information and enabling the comparison of different brain scans. The raw MRI images first need to go through a preprocessing pipeline. Compared to simply using the raw images as input to the model, we observed a significant increase in model performance when first applying this pipeline. The specific steps are skull stripping, voxel intensity standardization, registration, and sampling:

\begin{itemize}
\item Skull Stripping $-$ This step is to find the region of the image occupied by the brain, a process known as skull stripping. Once a mask for the brain in the image is identified, it is possible to remove the remaining body parts from the image (e.g. residual neck voxels). This is useful to decrease the amount of unnecessary information in the image, which act as noise, and also reduce the dimensionality, which may simplify the classification task~\cite{ref4,ref3}. The method employed is the Brain Extraction Tool \cite{Jenkinson2005} implemented in FSL \cite{Jenkinson2012}. In some of the early results, the method failed to correctly identify the brain area due to a large portion of the neck being present in the image. To address this issue, an option that employs segmentation steps for a more robust result was used instead.
\item Standardization $-$ This step is used to facilitate the comparison of different images by normalizing the voxel intensity values to a single standard range. Even images from the same person, taken in the same machine, and with the same sequence show different ranges of intensity values for the same brain areas \cite{NyulUdupa1998}. To address this issue we scale all images to a common value range. The input range is selected using the $0.2^{th}$ and $99.8^{th}$ percentiles to account for intensity outliers.
\item Registration $-$ This step aims to align the images based on brain structures to further aid in the comparison of different images. The registration is performed to the MNI152 atlas\footnote{http://nist.mni.mcgill.ca/?p=858} using affine transformation, which does not deform the images as it only includes translations, rotations, rescaling, and shearing. The similarity metric used was the correlation ratio. The FMRIB's Linear Image Registration Tool (FLIRT) \cite{Jenkinson2001} in FSL was used to perform the registration. 
\item Sampling $-$ The final step is to extract a patch of size $100\times100\times100$ voxels around the center. This region is known to be affected by the disease and was selected with the aid of an experienced radiologist. This region is shown in Figure~\ref{brain_region} superposed on the registration atlas.
\end{itemize}

\begin{table*}[ht]
\centering
\caption{A summary of the architectures used with the layers of each network indicated from top to bottom. Convolutional layers (Conv), either 2D or 3D according to the type of the network, are specified by the notation [filter side/stride, number of filters], with all filters being square/cubic and strides of 1 being omitted. All convolutional layers are followed by ReLU activation (not shown for brevity). Fully connected layers (FC) are followed by the number of units. All Max Pooling (Pool) layers were performed with a side and stride of 2.}\label{table:arch}
\begin{tabular}{llll}
\textbf{AlexNet 2D$+$C} & \textbf{AlexNet 3D} & \textbf{VGG16 2D$+$C} & \textbf{VGG16 3D} \\
\hline\\
Conv 11/4, 32 + Pool & Conv 11/4, 32 + Pool & (Conv 3, 32) x2   + Pool & (Conv 3, 16) x2 + Pool \\
Conv 5, 64 + Pool & Conv 5, 64 + Pool & (Conv 3, 64) x2 + Pool & (Conv 3, 32) x2 + Pool \\
Conv 3, 128 & Conv 3, 128 & (Conv 3, 64) x3 + Pool & (Conv 3, 32) x3 + Pool \\
Conv 3, 128 & Conv 3, 128 & (Conv 3, 128) x3 + Pool & (Conv 3, 64) x3 + Pool \\
Conv 3, 128 + Pool & Conv 3, 128 + Pool & (Conv 3, 128) x3 + Pool & (Conv 3, 64) x3 + Pool \\
FC 512 + ReLU + Dropout & FC 512 + ReLU + Dropout & FC 512 + ReLU + Dropout & FC 512 + ReLU + Dropout \\
FC 512 + ReLU + Dropout & FC 512 + ReLU + Dropout & FC 512 + ReLU + Dropout & FC 512 + ReLU + Dropout \\
FC 2 + Softmax & FC 2 + Softmax & FC 2 + Softmax & FC 2 + Softmax \\
\hline
\end{tabular}
\end{table*}

\subsection{Classification Models}
\label{sec:classification}

The second step is to fit CNN classification models to a training set of the data and then evaluate them on a separate test set. 

The design ideas for our CNN architectures are rooted in successful 2D natural image classification models, namely AlexNet~\cite{Krizhevsky@NIPS2012} and VGG16~\cite{Simonyan2014}. Table \ref{table:arch} describes the network architectures in detail. Specifically, we considered four different CNN architectures, two are variations of AlexNet \cite{Krizhevsky@NIPS2012} and the other two are variations of VGG16 \cite{Simonyan2014}. The ResNet~\cite{Korolev2017} architecture was not considered because authors in~\cite{exp1} showed the superiority of VGG16. 

The difference between each of the two AlexNet and VGG16 variations is how to handle the image dimensions. The first variation is a standard 3D CNN, which uses three-dimensional operations (convolutions and pooling). The second is an adapted 2D CNN, which we refer to as 2D $+$ Channel (2D$+$C). This variation represents one of the image dimensions using the layer depth, which usually is the color channel in RGB images. In effect, the model uses as input stacked 2D slices in a given plane, and has the advantage of using cheaper 2D operations and a lower number of parameters to fit. The plane represented as the depth and the number of slices to use are hyperparameters of this network. 

The age and sex of the subject were appended to the image representation before the fully connected layers in all architectures to aid the classifier~\cite{ref9}, given that these values are known to be related to the disease and easily accessible. 

All networks were trained for 50 epochs by minimizing the cross-entropy loss using the Adam optimizer \cite{Kingma@arXiv2014} with a fixed learning rate of 0.0001 and batches of 64 images.
The dropout rate was set to 0.5. All models were implemented using Keras with TensorFlow and trained on an Nvidia Tesla K80 GPU. After training, all models were calibrated in order to provide meaningful confidence values when being explained. Specifically, we used the Temperature Scaling \cite{Guo@arXiv2017} implemented with TensorFlow. A single parameter is used to scale the logits before applying the softmax function in the last layer of the network. The parameter is optimized by minimizing the same loss over the validation set while the remaining parameters remain fixed.

\subsection{Model Explanation}
\label{sec:explanation}

The last step is the novel explanation method which can be applied to a trained model to obtain a visual heatmap of the prediction for a given input brain scan.

Deep learning models involve sequences of non-linear convolutional layers and pooling that have different dimensionality from the input image, making it very difficult to interpret the relative importance of discriminating patterns in original data space. The Occlusion Test~\cite{Zeiler@ECCV2014} (OT) is a popular model explanation approach, in which the model is repeatedly tested with portions of the input image occluded to create a heatmap showing which parts of the brain image that actually have influence on the network output, that is, which parts of the brain image cause the model to change its outputs the most. Variations of the occlusion test are possible when the network parameters can be inspected directly~\cite{exp3}, leading to a number of approaches, including LRP~\cite{lrp}, DeepLift~\cite{deeplift}, CAM~\cite{cam}, and GradCAM~\cite{gradcam}.

As previously mentioned, while these approaches were shown to be effective in many applications, they may fail in explaining decisions about AD diagnosis due to the nature of the data. This may partially explain the limitations recently reported in~\cite{exp1,exp2}. To address these issues we propose the following alternative method.

\vspace{0.1in}
\noindent\textbf{The Swap Test $-$} In order to interpret the network predictions, we produce heatmaps to visualize the areas of the brain image most indicative of AD using a task-specific approach for explaining model decisions in the context of registered brain scans, which we refer to as the Swap Test (ST).

The Swap Test works as follows. 

\begin{itemize}
\item
Given a brain image $I$ to be explained, a reference image $R$ is randomly selected for comparison from a specific group. Brain images classified as CN are compared to true positives and images classified as AD are compared to true negatives. 
\item
A patch of a fixed size from $I$ is then copied to the same region in $R$, and the resulting image is used in the model to obtain a probability of AD. This is done to every cubic patch of the image and used to create a heatmap. 
\item
The process is then repeated for multiple references and averaged to account for individual variations in order to obtain a more robust result. This is only possible because the images were registered and the same brain structures occupy roughly the same positions in every image. 
\end{itemize}

With this approach, it is possible to find which regions of the image increase/decrease the predicted probability the most when the rest is known to be contributing to the opposite class. The patch size and number of references are hyperparameters of the method. There is a trade-off in choosing the patch size. Smaller patches increase the spatial resolution of the heatmap but they also increase the noise due to model uncertainty.

\section{Experiments}
\label{sec:experiments}

In this section, we discuss the data used in the experiments, our procedure for the classification evaluation and finally, our procedure for the explainability evaluation.

\subsection{Data}
\label{sec:data}

To fit the proposed model and validate its performance, data from two datasets were used. The first one is the Alzheimer's Disease Neuroimaging Initiative\footnote{http://adni.loni.usc.edu} (ADNI), which collects MRI scans from subjects, and clinical diagnosis of AD every 6 or 12 months. The second one is the Australian Imaging, Biomarker \& Lifestyle Flagship Study of Ageing\footnote{https://aibl.csiro.au/} (AIBL), which also collects both types of data. High-resolution 3D T1-weighted MR images were acquired on 3T scanners. We selected a single scan per subject visit, and when multiple sequences were available for a visit, we selected the most frequent one among the dataset distribution.

\subsection{Classification Evaluation}
\label{sec:evaluation}

In this section, we discuss how we evaluated the classification. All models are trained for separating the Alzheimer’s cohort (AD) from the cognitively normal cohort (CN). The initial set of images contained 826 subjects classified as CN and 422 classified as AD. Each image is a 3D tensor of intensity values with size $100\times100\times100$. 

Subjects were randomly divided into training, validation, and test sets, each of them containing respectively 1,779, 427, and 575 images (since subjects had multiple visits). We carefully included each subject into just one of the sets, in order to avoid biased generalization estimates due to same subject image similarities (that is, we ensured that every patient had scans in only one set). We chose the Area under the Receiver Operating Characteristic curve (or simply AUC) as our classification metric due to class imbalance~\cite{ref6}.

Results obtained with the VGGNet and ResNet architectures proposed in~\cite{exp1} were used for baseline comparison. Specifically, the VGGNet3D contains four blocks of 3D convolutional layers and 3D max pooling layers, followed by a fully connected layer, a batch normalization layer, a dropout layer, another fully connected layer, and the softmax output layer. The Adam optimizer was used to optimize model parameters. Learning rate was set to 0.000027, and the batch size was set to 5. The binary cross-entropy loss function was used.

For ResNet3D, a six-residual-block architecture is built. Each residual block consists of two 3D convolutional layers with $3\times3\times3$ filters that have a batch normalization layer and a RELU layer between them. Skip connections (identity mapping of a residual block) add a residual block element-by-element to the following residual block, explicitly enabling the following block to learn a residual mapping rather than a full mapping. For optimization, Nesterov accelerated stochastic gradient descent~\cite{nesterov} is used. Learning rate was set to 0.001, and batch size was set to 3. As with VGGNet3D, the cross-entropy loss function was used.

\subsection{Explainability Evaluation}

In this section, we discuss how we evaluated the explainability of our method.
Unlike the diagnosis classification for which we have the ground truth labels, we do not have the ground truth explanations, nor is it feasible to obtain. To evaluate the explanations we then adopt the methodology proposed by Montavon et al.~\cite{Montavon@DSP2018}, which defines two desirable properties of explanations: continuity and selectivity. Next, we provide a brief explanation of each and how they were implemented.

\begin{itemize}
\item Continuity: If this property is satisfied we expect similar images to have similar explanations. If two brain images are nearly equivalent, then the explanations of their predictions should also be nearly equivalent. In other words, we expect the explanation function to be continuous. 

Given an input image, we can obtain a similar image by applying a small perturbation. We measure continuity by calculating the $\ell2$ norm between the original heatmap and the heatmap derived from the perturbed image, as shown in the following equation:

\begin{equation}
\max\frac{||C(x) - C(x')||_1}{||x - x'||_2}
\nonumber
\end{equation}

\noindent where $x$ is the original image, $x'$ is the modified image, and $C(x)$ and $C(x')$ are the corresponding heatmaps. We use the expected norm between heatmaps from different images as a reference for comparison. If there is continuity, we expect the measured value to be significantly lower than the reference value.

\item Selectivity: This property states that the most relevant regions in the heatmap should have a higher impact on the model if removed. Selectivity quantifies how fast the class probability changes when removing the most relevant features. 

We measure selectivity with a reversed version of both swap and occlusion tests. Instead of swapping/occluding each cubic region, the rest of the image is swapped/occluded to measure its separate relevance. 

We then measure the Pearson correlation coefficient ($\rho$) between the standard and reversed heatmaps, and thus selectivity values are in the range of $-$1 to $+$1 (perfect positive correlation). If there is selectivity, we expect a negative correlation since regions of higher relevance in the standard heatmap would cause higher drops to the prediction in the reversed heatmap when removed.
\end{itemize}

\section{Results}
\label{sec:results}

We start this section by reporting the classification performance of all four CNN architectures. We then report the explainability performance, starting with the continuity and selectivity values observed for all architectures, followed by a visual qualitative evaluation of the heatmaps produced by the proposed Swap Test in identifying key brain regions that are known to be associated with AD.

\subsection{Classification Performance}
We first conducted preliminary experiments to select the hyperparameters of the 2D+C networks. We observed that the sagittal plane provided slightly better classification results, although by a small margin. We also performed multiple experiments with a different number of slices, and we observed no significant difference in performance when using up to 1 every 5 slices. We adopted these values for the rest of the experiments.

\begin{figure}
\centering
\includegraphics[width=\linewidth]{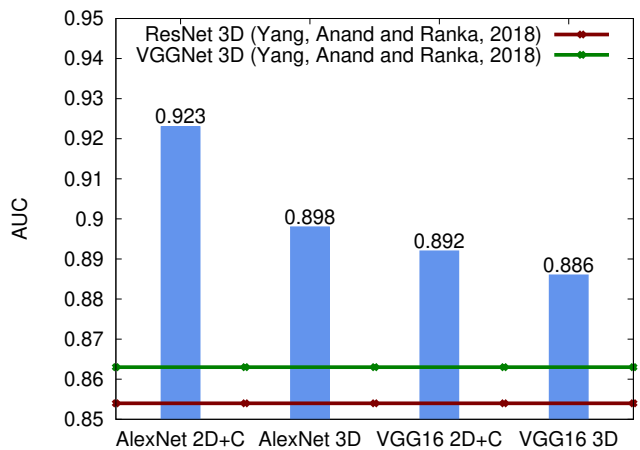}
\caption{Classification performance of different models.}
\label{fig:comparison}
\end{figure}

Figure~\ref{fig:comparison} shows the final classification performance for each model. Results for the baseline models are also shown in the figure. Clearly, all models achieved similar AUC values, except for the AlexNet 2D$+$C which outperformed the other models by a larger margin, despite having a smaller capacity and not performing 3D operations. Gains in classification performance range from 2.4\% (VGG16 3D vs. VGGNet 3D) to 8.1\% (AlexNet 2D$+$C vs ResNet 3D). Overall, we conclude that both the 3D and 2D$+$C networks showed their effectiveness, with no clear winner. Finally, regarding the different CNN architectures, we conclude that AlexNet performs better than VGGNet, and VGGNet performs better than ResNet.

\begin{figure*}[h]
\centering
\includegraphics[width=\linewidth]{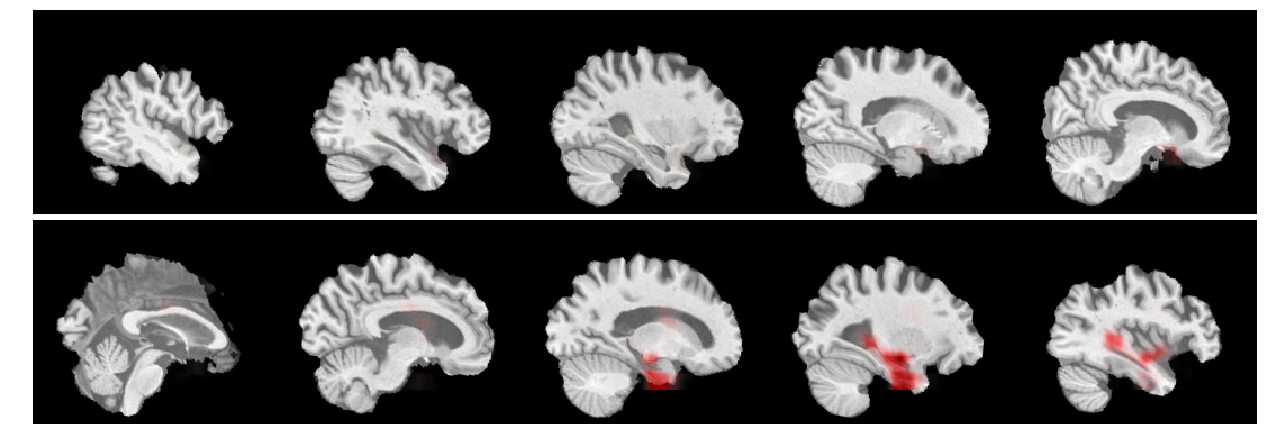}
\caption{Visual assessment (best viewed in color): ST explanation for a sample true positive image shown in ten equally spaced sagittal slices displayed from left to right and top to bottom. Here the model focused on the left hippocampus and ventricles.}
\label{fig:st_ex}
\end{figure*}

\subsection{Explainability Performance}

Table~\ref{table:metrics} shows the continuity and selectivity values for each model. Results for both metrics consist of the average over 50 randomly sampled images. Lower values for continuity indicate that similar images have more similar heatmaps. The baseline continuity value is 30.596, which was calculated as the average $\ell2$ norm between all the heatmaps from different images we generated. Both ST and OT heatmaps were generated with patches of 20 voxel wide cubes, which we found to be a good compromise between spatial resolution and noise. We used 5 reference images for ST, which we found to be enough to produce robust heatmaps.

\begin{table}[ht]
\centering
\caption{Explainability performance of both Swap Test and Occlusion Test measured in four different CNN architectures. Lower values are better for both metrics.}
\label{table:metrics}
\begin{tabular}{lrrrr}
               & \textbf{Swap} & \textbf{Occlusion} & \textbf{Swap} & \textbf{Occlusion}\\
\textbf{Model} & \textbf{Continuity} & \textbf{Continuity} & \textbf{Selectivity} & \textbf{Selectivity}\\
\hline\\
AlexNet 2D$+$C & \textbf{16.861} & 30.361 & \textbf{-0.615} & -0.059 \\
VGG16 2D$+$C   & \textbf{17.431} & 24.928 & \textbf{-0.643} & 0.224 \\
AlexNet 3D     & \textbf{16.035} & 37.887 & \textbf{-0.364} & 0.215 \\
VGG16 3D       & \textbf{15.492} & 41.879 & \textbf{-0.485} & 0.039 \\
\hline
\end{tabular}
\end{table}

The continuity values for ST in all models were significantly lower than the baseline value, as required by continuity, which was not observed for OT. This indicates that similar images according to OT may actually generate completely unrelated heatmaps, implying the absence of continuity. Further, there is a strong negative correlation of the selectivity values for ST for all models, as we hypothesized. For OT, on the other hand, there was either no significant correlation or a small positive correlation.

\subsection{Qualitative Analysis of Visual Explanations}

To visually check the quality of the heatmaps generated by our proposed visual explanation approach, we take one MRI scan for visual inspection and present the corresponding heatmaps. Figure~\ref{fig:st_ex} shows the ST explanation for a sample image of an AD patient that was correctly classified using AlexNet 2D$+$C. The 3D image is shown in sagittal slices displayed from left to right and top to bottom. These heatmaps should be read as an indication of where the network model sees evidence of AD. As can be seen, the model uses meaningful patterns as a basis for its decision, since the heatmaps correlate with what is known from the literature. Specifically, the network assigned importance the left hippocampus and ventricles.

\begin{figure}
\centering
\includegraphics[width=\linewidth]{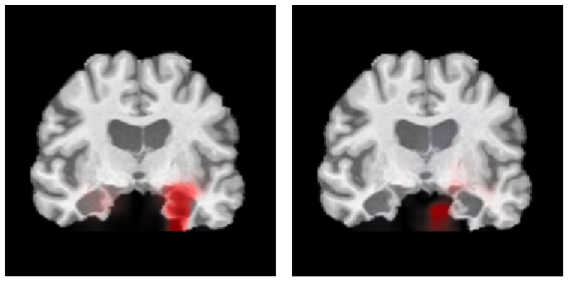}
\caption{Visual assessment (best viewed in color): ST explanation (left) and OT explanation (right) for a sample true positive image shown in coronal slices. While the ST correctly highlighted the hippocampus, the OT highlighted a region outside the brain.}
\label{fig:st_ex2}
\end{figure}

For comparison, figure~\ref{fig:st_ex2} shows both the ST and OT explanations for another sample true positive image, shown in coronal slices. The ST correctly highlighted the region of the hippocampus, with more emphasis given to the left side. On the other hand, the OT identified a region outside of the brain as being relevant.

\section{Conclusion}
\label{sec:conclusion}

Alzheimer's Disease is the most common form of dementia and currently there is no available cure for the disease. The available AD diagnostics that use deep learning models function as black boxes, meaning that results do not include any explanation of why the machine decides a patient has the disease or not. While deep learning models are extraordinarily accurate, their adoption in healthcare has been slow because doctors and regulators cannot verify their results. It has become clear that the ability to provide understandable and reliable explanations is a crucial aspect of the implementation of any automated diagnosis model for AD, in order to convey trust when adopted and to comply with upcoming regulations. 

In this paper, we proposed a novel model explanation approach specifically designed for registered brain MRI scans. Our experiments showed that the proposed approach exhibits desirable explainability properties, while the opposite trend was observed when using a typical occlusion test. Finally, our results indicate that 2D$+$C models provide better selectivity numbers, while 3D models provide slightly better continuity numbers. Further, by visually inspecting the results we found that our proposed Swap Test explanation approach indicates regions that are classically related to AD diagnosis.

However, we still cannot make any claim about causal relationships, and future studies are necessary to more systematically investigate the relationship between manifested neurobiological markers and Swap Test explanations.

\bibliographystyle{IEEEtran}
\bibliography{refs}

\end{document}